\documentclass[letterpaper]{article} 
\usepackage[preprint]{aaai2027} 
\usepackage[hyphens]{url}  
\usepackage{graphicx} 
\usepackage{natbib}  
\usepackage{caption} 
\usepackage{amsmath}
\usepackage{amssymb}
\usepackage{multirow}
\usepackage{algorithm}
\usepackage{algorithmic}
\usepackage{booktabs}
\usepackage{newfloat}
\usepackage{listings}
\DeclareCaptionStyle{ruled}{labelfont=normalfont,labelsep=colon,strut=off} 
\floatstyle{ruled}
\newfloat{listing}{tb}{lst}{}
\floatname{listing}{Listing}

\usepackage{booktabs}

\title{JEPA-WAM: Connecting Generated Visual Instructions to\\ World Action Models through JEPA Latent Representations}

\author{
Tianbin Liu\textsuperscript{\rm 1}, Jian Zhu%
\textsuperscript{\rm 1}%
\textsuperscript{,}%
\thanks{Project lead.}%
\textsuperscript{,}%
\corresponding%
, Taiyi Su\textsuperscript{\rm 1}, Jianjun Zhang\textsuperscript{\rm 1, 2}, Chong Ma\textsuperscript{\rm 1, 2}, Zitai Huang\textsuperscript{\rm 1, 2}, Yi Xu\textsuperscript{\rm 1}
\textsuperscript{,}%
\corresponding
}

\affiliations{
\textsuperscript{\rm 1}AIRC, Midea Group \textsuperscript{\rm 2}Tongji University \\
liutb27@midea.com, jianzhu823@gmail.com, xuyi42@midea.com}

\begin{document}

\maketitle

\begin{abstract}
World Action Models (WAMs) have demonstrated strong robotic manipulation capabilities by augmenting pretrained video generative models with action experts. However, current WAMs still show limited instruction-following ability when conditioned solely on text instructions. We argue that this limitation stems in part from a structural imbalance in robot-learning data: rich visual-action trajectories are often paired with sparse and repetitive language annotations, allowing policies to identify tasks from visual context and motion regularities rather than grounding the instruction itself. To address this limitation, we introduce \textbf{JEPA-WAM}, which augments each text instruction with a bank of stochastically generated visual instructions, providing diverse visual cues for instruction following. Specifically, JEPA-WAM uses an off-the-shelf text-to-image generator to sample multiple task-completion images conditioned on the text instruction, without training the generator. Although these generated images may differ from the current visual scene in appearance and layout, they remain semantically aligned with the instruction and serve as visual goal references. To focus on task-level semantics beyond appearance, we encode these references with a frozen V-JEPA 2.1 encoder. The resulting dense goal representations are compressed into compact goal tokens that condition both the video and action experts through cross-attention. We further construct a real-robot instruction-following benchmark covering in-distribution, out-of-distribution scene, and out-of-distribution instruction settings. On this benchmark, JEPA-WAM achieves success rates of 87.3\%,
74.5\%, and 80.9\% in these three settings, outperforming
$\pi_0$ and Fast-WAM by at least 10.0, 27.3, and 14.5 percentage
points, respectively.
\end{abstract}


\section{Introduction}
\label{sec:introduction}

Recent World Action Models (WAMs) adapt pretrained video generative models for robot control by coupling them with action experts, jointly learning future visual states and robot actions \cite{yuan2026fastwam,liang2025videogenerators,bi2026motus, ye2026worldactionmodelszeroshot}. By transferring visual dynamics learned from large-scale video pretraining, these models acquire rich physical priors and demonstrate strong manipulation performance in both simulated and real-world environments.

However, strong manipulation performance does not necessarily imply faithful instruction following. Existing language-conditioned WAMs \cite{yuan2026fastwam,liang2025videogenerators,bi2026motus, ye2026worldactionmodelszeroshot} typically use a single trajectory-level text instruction to condition visual dynamics and action prediction. Yet this language signal is static and substantially less diverse than the temporally dense visual-action supervision available throughout the trajectory \cite{xiao2023robotic}. This modality imbalance can allow WAMs to rely on visual context and dataset-specific regularities rather than faithfully grounding the instruction \cite{xing2025shortcut,fang2026vision,xu2026apt}. Such shortcuts may work well on familiar task distributions but leave the policy vulnerable to scene shifts and unseen instruction formulations.

Building on this observation, we seek to enrich each trajectory-level instruction with additional task-semantic supervision in a form that can effectively condition WAM control. Visual instructions can make target objects, spatial relations, and desired outcomes more explicit than a single repeated language annotation. We therefore augment each text instruction with multiple visual hypotheses of task completion. Because visual instructions are intended primarily to preserve semantic consistency with the text instruction while providing diverse realizations of the task, we use an off-the-shelf text-to-image model, without fine-tuning it on the robot dataset, to generate these visual hypotheses \cite{rombach2022highresolution,saharia2022photorealistic, feng2026dreamlite}. This choice, however, introduces a representation challenge. Since the generated images are conditioned only on text rather than the current observation, they may differ from the robot's actual workspace in appearance, spatial layout, viewpoint, and other scene-specific details. Directly conditioning the policy on raw generated pixels would therefore entangle task-relevant semantics with nuisance visual variation. We consequently require a representation that abstracts away low-level appearance differences while preserving the spatially structured semantics needed to identify target objects, relations, and task outcomes. V-JEPA 2.1 is particularly suited to this role because it learns predictive representations in latent space rather than reconstructing raw pixels, while its dense features retain explicit spatial structure and semantic coherence \cite{bardes2024vjepa,murlabadia2026vjepa21}. We therefore use a frozen V-JEPA 2.1 encoder to transform generated visual instructions into robust latent representations that can effectively condition WAM control.

To realize this approach, we introduce \textbf{JEPA-WAM}, which connects generated visual instructions to robot control through V-JEPA latent representations. For each text instruction, we construct an offline bank of task-completion images generated with different random seeds. During training, one image is sampled from the corresponding bank and encoded by the frozen V-JEPA 2.1 encoder. Its dense features are spatially pooled, projected into the WAM conditioning space, and provided to both the video and action experts through cross-attention. This design allows the generated task semantics to condition both visual dynamics modeling and action generation through compact JEPA latent representations. To evaluate instruction-following performance and robustness, we further construct a real-robot benchmark containing 11 behaviorally overlapping manipulation tasks. These tasks share objects and action primitives but require distinct outcomes, making the instruction essential for identifying the intended behavior. We evaluate the policies under three settings: an in-distribution (ID) setting using familiar scenes and training instructions, an out-of-distribution scene (OOD-S) setting introducing familiar and unseen distractors, and an out-of-distribution instruction (OOD-I) setting using previously unseen instruction reformulations, including formulations that resemble instructions for behaviorally related tasks. JEPA-WAM achieves success rates of 87.3\%, 74.5\%, and 80.9\% under ID, OOD-S, and OOD-I, respectively, exceeding $\pi_0$ and Fast-WAM by at least 10.0, 27.3, and 14.5 percentage points.

Our contributions are threefold: 
\begin{itemize} 
    \item We introduce a scalable visual-instruction augmentation strategy that enriches trajectory-level text instructions with semantically aligned and visually diverse task-completion images generated by an off-the-shelf text-to-image model. 
    \item We propose JEPA-WAM, which uses a frozen V-JEPA 2.1 encoder to transform generated visual instructions into spatially structured latent representations robust to scene-specific appearance variation. 
    \item We construct an 11-task real-robot benchmark with behaviorally overlapping tasks and evaluate instruction following under in-distribution, out-of-distribution scene, and out-of-distribution instruction settings. JEPA-WAM consistently outperforms representative WAM and VLA baselines.
\end{itemize}

\section{Related Work}

\begin{figure*}[t]
    \centering
    \includegraphics[width=\textwidth]{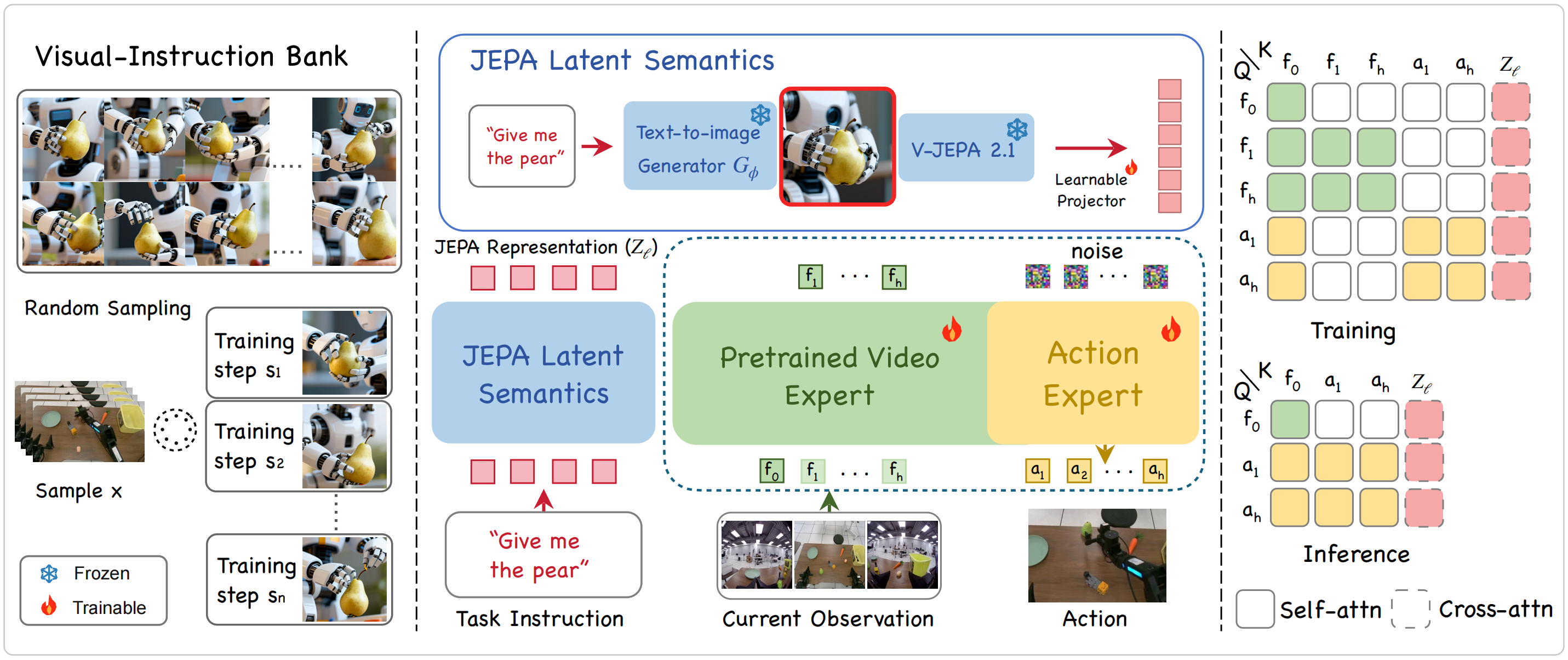}
\caption{
\textbf{Overview of JEPA-WAM.}
For each language instruction $\ell$, we construct a visual-instruction bank
$\mathcal{B}_{\ell}$ containing $K$ images sampled from an
instruction-conditioned text-to-image generator.
One visual instruction is randomly
sampled and encoded by frozen V-JEPA~2.1 and a learnable projector into compact
representations $Z_{\ell}$ at each training step.
These representations condition the video and action experts together with the
task instruction and current observations.
The right panel shows the self- and cross-attention masks used during training
and inference.
}
\label{fig:jepa-wam-overview}
\end{figure*}

\subsection{World Action Models}

Vision-language-action models condition action prediction on visual
observations and language instructions, providing a general interface for
language-guided robot control \cite{zitkovich2023rt,kim2024openvla,black2024pi_0}. World Action Models
(WAMs) augment action learning with objectives that model future visual states,
thereby coupling robot control with learned visual dynamics
\cite{liang2025videogenerators,ye2026worldactionmodelszeroshot,
bi2026motus,yuan2026fastwam}. Video Policy combines video generation with
action decoding, while DreamZero jointly models future video and robot actions
\cite{liang2025videogenerators,ye2026worldactionmodelszeroshot}. Motus
integrates understanding, video-generation, and action experts within a unified
architecture, whereas Fast-WAM retains video co-training but removes explicit
future generation at test time
\cite{bi2026motus,yuan2026fastwam}. DSWAM further combines a WAM executor with
an optional vision-language planner and studies WAM and VLA execution under a
matched-data setting \cite{zhu2026dswam}.

However, existing WAM evaluations have primarily emphasized task execution and
physical generalization, while instruction grounding under behaviorally
overlapping tasks remains comparatively underexplored. JEPA-WAM studies this
problem through controlled scene and instruction shifts designed to expose
reliance on visual and linguistic shortcuts.

\subsection{Instruction Following and Visual Instructions}

Language-conditioned robot policies are commonly trained on demonstrations
paired with task descriptions. Because language annotations are expensive and
often less diverse than the corresponding visual-action trajectories, prior
work has explored automatic instruction augmentation. DIAL uses pretrained
vision-language models to propagate language labels to unlabeled robot
demonstrations \cite{xiao2023robotic}. More recent studies show that limited
within-dataset diversity can induce shortcut learning in generalist robot
policies, causing them to rely on task-irrelevant visual regularities
\cite{xing2025shortcut}. Counterfactual evaluations further demonstrate that
VLA policies may default to visually familiar behaviors despite changes in the
language instruction \cite{fang2026vision}.

Visual goals provide an alternative interface for specifying desired behavior.
GRIF aligns language with representations of goal-state changes, while
Interleave-VLA supports interleaved image-text task specifications
\cite{myers2023goal,fan2025interleave}. Text-to-image models offer another
source of visual task information by generating semantically aligned images
from language \cite{rombach2022highresolution,saharia2022photorealistic}.
JEPA-WAM uses this capability to construct instruction-specific visual banks
without requiring user-provided goal images, additional robot trajectories, or
joint training of the image generator.

\subsection{Predictive Visual Representation Learning}

Joint-Embedding Predictive Architectures learn representations by predicting
masked content in latent space rather than reconstructing raw pixels. I-JEPA
shows that representation-space prediction can produce semantic image features
without relying on pixel-level reconstruction
\cite{assran2023self}. V-JEPA extends this principle to video and
learns transferable representations for both appearance- and motion-sensitive
tasks \cite{bardes2024vjepa}. V-JEPA 2.1 further improves dense visual
representations through dense predictive learning and deep self-supervision,
producing features with explicit spatial structure and semantic coherence
\cite{murlabadia2026vjepa21}. Recent analysis also suggests that
latent-prediction video models degrade more gracefully under pixel corruptions
and occlusions than reconstruction-based alternatives
\cite{alrasheed2026latent}.

These properties motivate our use of V-JEPA 2.1 for generated images whose
appearance may differ from the robot workspace. JEPA-WAM does not train the
V-JEPA predictor or introduce an additional predictive objective. Instead, it reuses the
frozen encoder's dense features as conditioning for the WAM.

\section{JEPA-WAM}
\label{sec:method}

JEPA-WAM augments language-conditioned world-action modeling with diverse
visual hypotheses of task completion. The method consists of two main
components. First, generated visual-instruction banks expand each language
instruction into multiple visual realizations and enable stochastic
instruction sampling during training. Second, the sampled images are
transformed into spatially structured latent visual semantics and incorporated
into both visual dynamics modeling and action prediction.
Fig.~\ref{fig:jepa-wam-overview} provides an overview of the complete
architecture and information flow.

\subsection{Generated Visual-Instruction Banks}
\label{sec:visual_instruction_bank}

A single generated image captures only one possible realization of task
completion and may contain incidental appearance or layout patterns. Rather
than treating one generated image as a deterministic visual goal, we associate
each language instruction with a bank of diverse task-completion hypotheses.

Let $\ell$ denote a language instruction and let $G_{\phi}$ be a pretrained
text-to-image generator. For each unique instruction, we construct an offline
visual-instruction bank
\begin{equation}
\mathcal{B}_{\ell}
=
\left\{
g_{\ell}^{(k)}
=
G_{\phi}(\ell;\epsilon_k)
\right\}_{k=1}^{K},
\label{eq:visual_bank}
\end{equation}
where $\epsilon_k$ is a random sampling seed and $K$ is the bank size.
Different seeds produce varied appearances, compositions, and spatial layouts
under the same language condition. The resulting bank forms a finite
approximation of the generator-induced conditional distribution
$p_{\phi}(g\mid\ell)$.

The generator is used only for offline bank construction and is neither
fine-tuned nor jointly optimized with the robot policy. Because generation is
conditioned on the language instruction rather than the current observation,
the resulting images represent task-level completion hypotheses rather than
pixel-accurate predictions of the future robot scene.

During training, each trajectory segment associated with instruction $\ell$
is paired with an image uniformly sampled from $\mathcal{B}_{\ell}$. This
stochastic pairing exposes the policy to multiple visual realizations of the
same task semantics without requiring additional robot demonstrations or
manually constructed language paraphrases.

\subsection{Latent Visual Semantics for World-Action Modeling}
\label{sec:jepa_conditioning}

The visual-instruction banks provide diverse task-completion hypotheses, but
their raw pixels may differ substantially from the current robot workspace in
appearance and layout. Effective visual instructions should therefore suppress
such scene-specific variation while preserving the spatial semantics needed to
identify task-relevant objects, relations, and outcomes. JEPA-WAM achieves this
through a frozen V-JEPA encoder and integrates the resulting latent visual
semantics into both the video and action experts.

\paragraph{Latent visual semantics.}
Let $E_{\mathrm J}$ denote a frozen V-JEPA~2.1 encoder and let $P_{\psi}$
denote a learnable projector. We use V-JEPA~2.1 because its dense latent features retain spatial
structure and semantic coherence without requiring pixel-level
reconstruction. To accommodate the temporal tubelet size of two used
by V-JEPA~2.1, we duplicate each static visual instruction
to form a two-frame clip. For each generated visual instruction
$g_{\ell}^{(k)}$, we compute
\begin{equation}
Z_{\ell}^{(k)}
=
P_{\psi}
\left(
\operatorname{Pool}_{8\times 8}
\left(
E_J
\left(
\left[
g_{\ell}^{(k)},\, g_{\ell}^{(k)}
\right]
\right)
\right)
\right)
\in
\mathbb{R}^{64\times d_c},
\label{eq:jepa_tokens}
\end{equation}
where $\operatorname{Pool}_{8 \times 8}$ spatially compresses the dense
V-JEPA features into 64 tokens and $P_{\psi}$ maps them into the WAM
representation dimension $d_c$. Spatial pooling reduces the conditioning cost
while retaining coarse spatial organization. The V-JEPA encoder remains
frozen, whereas the projector is optimized jointly with the robot policy.

\begin{figure}[t]
    \centering
    \includegraphics[width=\columnwidth]{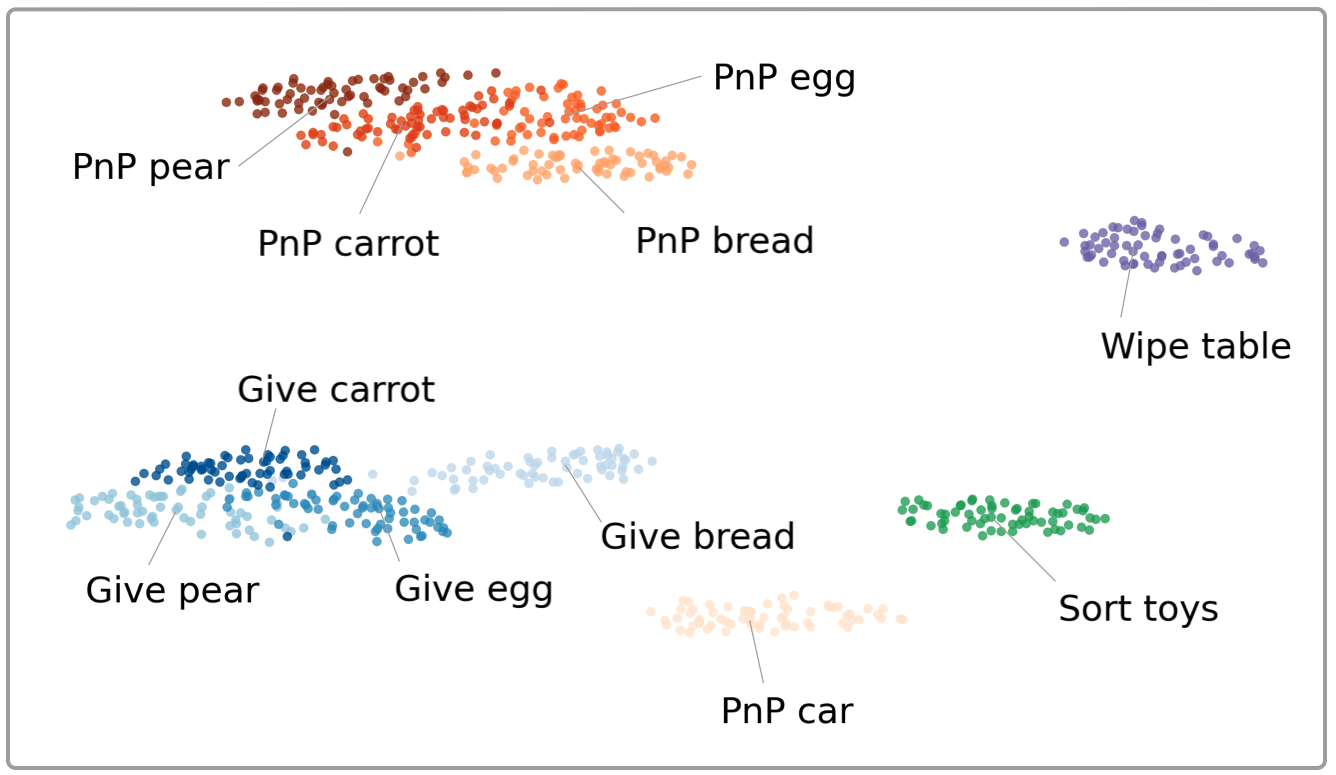}
\caption{
UMAP~\cite{mcinnes2018umap} visualization of frozen V-JEPA 2.1 representations for 704 generated
images from the 11 benchmark tasks defined in Sec.~\ref{sec:benchmark}. Each point denotes one image. Dense
patch representations are globally averaged and L2-normalized before
projection. The embedding reveals task-level structure among generated
visual hypotheses. Related behaviors form coherent
regions, while distinct interaction types remain separated.
}
\label{fig:vjepa_umap}
\end{figure}

Encoding the $K$ images in $\mathcal{B}_{\ell}$ induces an empirical latent
distribution
\begin{equation}
\widehat p_K(Z\mid\ell)
=
\frac{1}{K}
\sum_{k=1}^{K}
\delta\!\left(Z-Z_{\ell}^{(k)}\right),
\label{eq:latent_distribution}
\end{equation}
where $\delta(\cdot)$ denotes the Dirac delta function. This formulation
captures that JEPA-WAM is trained with stochastic samples from a distribution
of latent task hypotheses rather than with a single fixed visual reference.
Fig.~\ref{fig:vjepa_umap} provides a qualitative visualization of the
task-level structure exhibited by these representations.

\paragraph{World-action modeling.}
The intended task should guide both how the visual scene is expected to evolve
and which robot actions should be executed. We therefore provide the same
latent visual semantics to both the video and action experts.

Let $C_{\ell} \in \mathbb{R}^{L \times d_c}$ denote the language
conditioning sequence, and let $P_s(s) \in \mathbb{R}^{1 \times d_c}$ denote
the projected token of the current robot state $s$. The complete conditioning
sequence is
\begin{equation}
C
=
\left[
C_{\ell};
P_s(s);
Z_{\ell}
\right].
\label{eq:conditioning_sequence}
\end{equation}
The video expert uses $C$ to model task-guided visual dynamics, while the
action expert uses the same sequence to predict the denoising direction of the
action trajectory. We otherwise retain the directional masked-attention
structure of Fast-WAM. Consequently, the latent visual semantics influence
action generation directly through the action expert and indirectly through
the conditioned video representations.

\paragraph{Training and inference.}
Let $\tau$ denote a robot demonstration associated with instruction $\ell$.
At each training step, we uniformly sample an index
$k\in\{1,\ldots,K\}$ from the corresponding visual-instruction bank. The
training objective is
\begin{equation}
\min_{\theta,\psi}
\mathbb{E}_{(\tau,\ell)\sim\mathcal D}
\mathbb{E}_{k\sim\operatorname{Uniform}(\{1,\ldots,K\})}
\left[
\mathcal L_{\mathrm{WAM}}
\left(\theta;\tau,\ell,Z_{\ell}^{(k)}\right)
\right],
\label{eq:jepawam_objective}
\end{equation}
where $\theta$ denotes the trainable WAM parameters and $\psi$ denotes the
learnable projector. We retain the original joint flow-matching objective
\begin{equation}
\mathcal L_{\mathrm{WAM}}
=
\lambda_v \mathcal L_{\mathrm{video}}
+
\lambda_a \mathcal L_{\mathrm{action}},
\label{eq:wam_objective}
\end{equation}
where $\mathcal L_{\mathrm{video}}$ and $\mathcal L_{\mathrm{action}}$
denote the video-latent and action-prediction losses, respectively, and
$\lambda_v$ and $\lambda_a$ are scalar weighting coefficients that balance
their relative contributions to the joint training objective. Only
$\theta$ and $\psi$ are optimized; the text-to-image generator $G_{\phi}$
and V-JEPA encoder $E_{\mathrm J}$ remain fixed.

At deployment, a visual instruction is generated and encoded when a language
instruction is received or changed. Its latent representation is then reused
across subsequent control steps, avoiding repeated image generation and JEPA
encoding for every action query.

\paragraph{Implementation details.}
In our implementation, we instantiate the image generator $G_{\phi}$ with the
compact 0.39B DreamLite model~\cite{feng2026dreamlite} and the visual encoder
$E_{\mathrm J}$ with the 80M-parameter V-JEPA~2.1 ViT-B/16
checkpoint~\cite{murlabadia2026vjepa21}. We select these lightweight
variants to keep the complete conditioning pipeline on a single device and
eliminate inter-GPU communication during closed-loop control. The complete
inference stack, including image generation and JEPA encoding, runs on a single
NVIDIA RTX 5090. DreamLite uses a fixed prompt template across all tasks, with
only the task instruction and sampling seed varied. Unless otherwise specified,
the main JEPA-WAM model uses a bank of 1,024 generated images per instruction.
The exact generation prompt is provided in
Supplementary Sec.~A.

\begin{figure*}[!t]
    \centering
    \includegraphics[width=\textwidth]{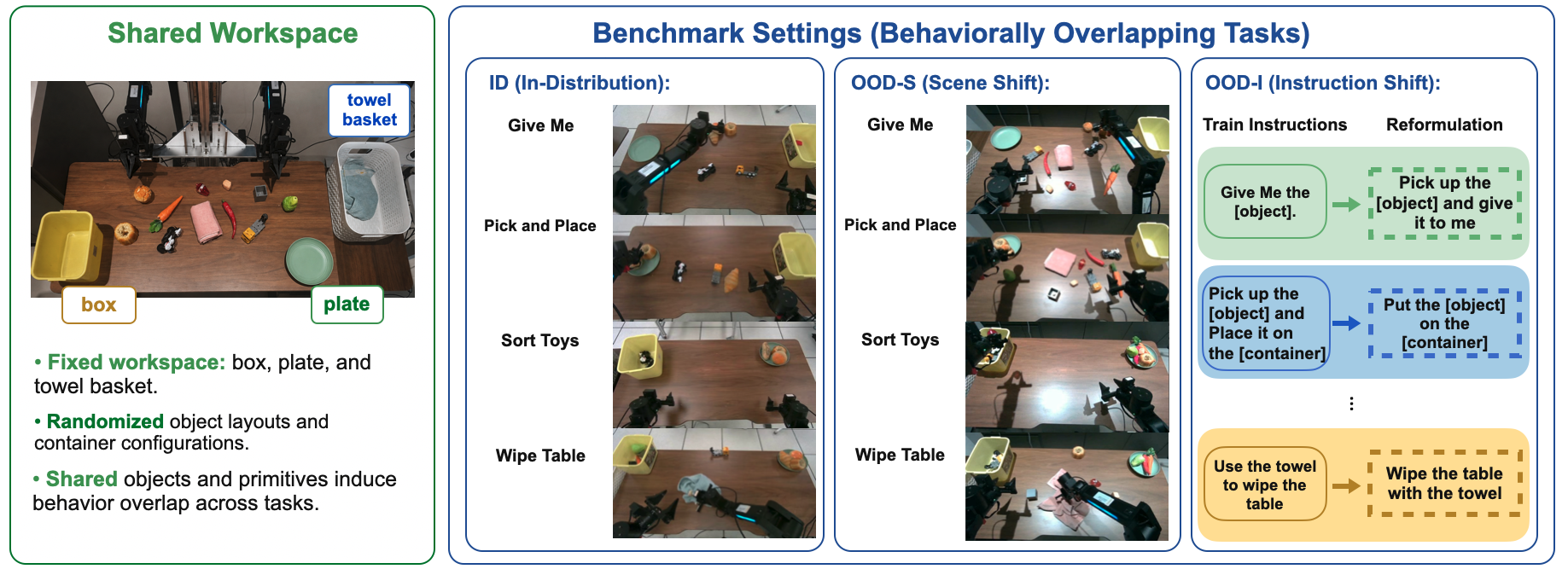}
\caption{
Overview of the real-robot instruction-following benchmark.
The benchmark uses a shared workspace with fixed containers
(box, plate, and towel basket) while randomizing object layouts and
container configurations across trials. It evaluates 11 behaviorally
overlapping tasks under three complementary settings: in-distribution
conditions (ID), scene shifts (OOD-S), and instruction shifts (OOD-I).
ID follows the training distribution, OOD-S introduces new scene
variations with the original instructions, whereas OOD-I changes only
the instruction formulations while keeping the scenes and intended
behaviors unchanged. Representative OOD-I examples illustrate the
correspondence between training instructions and unseen reformulations.
}
    \label{fig:benchmark}
\end{figure*}

\section{Robot Instruction-Following Benchmark}
\label{sec:benchmark}

We construct a real-robot benchmark to evaluate whether a policy follows the
specified instruction when visual context alone is insufficient to determine
the intended behavior. The benchmark contains 11 behaviorally overlapping
tabletop tasks that reuse objects, containers, workspace layouts, and action
primitives while requiring different outcomes. This design makes the language
instruction necessary for disambiguating both the target object and the
requested behavior. We evaluate instruction following under in-distribution
conditions, scene shifts, and instruction reformulations.
Fig.~\ref{fig:benchmark} summarizes the task suite and evaluation settings.

\subsection{Task Suite and Data Distribution}
\label{sec:benchmark_tasks}

The demonstration dataset used to train all evaluated policies contains four
\textit{Give Me} tasks, five \textit{Pick-and-Place} (PnP) tasks, one
\textit{Sort Toys} task, and one \textit{Wipe Table} task. \textit{Give Me} requires
moving the instructed object to the front handover boundary, whereas
\textit{PnP} requires placing a toy car in the box or a food object on the
plate. \textit{Sort Toys} requires assigning cars to the box and bread items
to the plate, while \textit{Wipe Table} requires retrieving a towel, wiping the
exposed tabletop without disturbing protected objects, and returning
the towel to its basket. Complete scene compositions are provided in
Supplementary Sec.~B.

All policies are trained on the same set of 1,370 real-robot demonstrations.
Each PnP task contains 150 demonstrations, as do \textit{Sort Toys} and
\textit{Wipe Table}, whereas each \textit{Give Me} task contains 80
demonstrations. This controlled frequency imbalance creates a stronger training
prior toward PnP behaviors and tests whether a policy can still execute the
less frequent handover behavior when instructed. Across demonstrations, the box and
plate sides are balanced, and all remaining object positions are
randomized.

\subsection{Evaluation Settings}
\label{sec:benchmark_settings}

The ID setting uses familiar scenes and training instructions, whereas
OOD-S and OOD-I evaluate robustness to scene and instruction shifts,
respectively.

\paragraph{ID.}
Policies are evaluated using the training instruction formulations in
scenes sampled from the corresponding training distribution.

\paragraph{OOD-S.}
The training instructions are retained, while the workspace is made
more ambiguous through additional familiar and unseen distractors.
For Give Me and PnP, policies must select the instructed object and
execute the requested behavior among multiple plausible alternatives.
Sort Toys additionally tests whether the demonstrated food-versus-
non-food sorting rule transfers to categories absent from its training
demonstrations, while Wipe Table introduces familiar, unseen, or mixed
removable clutter. Full scene configurations are provided in
Supplementary Sec.~B.

\paragraph{OOD-I.}
The scene and intended task remain unchanged, but the training
instruction is replaced with a semantically equivalent, previously
unseen formulation. Some reformulations deliberately overlap
lexically and structurally with instructions for behaviorally related
tasks. For example, ``give me the egg'' becomes ``pick up the egg and
pass it to me,'' which resembles the PnP instruction while specifying
a different outcome. One fixed OOD-I instruction is defined for each
task and used for all policies and trials; the complete list is
provided in Supplementary Sec.~B.

\begin{table*}[!t]
\centering

\begingroup
\setlength{\tabcolsep}{1mm}

\begin{tabular*}{\textwidth}{
    @{\extracolsep{\fill}}
    l
    @{\hspace{5mm}}c@{\hspace{6mm}}c@{\hspace{1mm}}c
    @{\hspace{5mm}}c@{\hspace{6mm}}c@{\hspace{1mm}}c
    @{\hspace{5mm}}c@{\hspace{6mm}}c@{\hspace{1mm}}c
}
\toprule
& \multicolumn{3}{c}{ID}
& \multicolumn{3}{c}{OOD-S}
& \multicolumn{3}{c}{OOD-I} \\
\cmidrule(lr){2-4}
\cmidrule(lr){5-7}
\cmidrule(lr){8-10}
Task
& $\pi_0$ & Fast-WAM & \textbf{JEPA-WAM}
& $\pi_0$ & Fast-WAM & \textbf{JEPA-WAM}
& $\pi_0$ & Fast-WAM & \textbf{JEPA-WAM} \\
\midrule

Give bread
& 30.0 & 60.0 & 100.0
& 20.0 & 50.0 & 90.0
& 20.0 & 60.0 & 100.0 \\

Give pear
& 10.0 & 70.0 & 60.0
& 0.0 & 60.0 & 50.0
& 10.0 & 60.0 & 90.0 \\

Give egg
& 10.0 & 50.0 & 80.0
& 10.0 & 50.0 & 50.0
& 0.0 & 30.0 & 60.0 \\

Give carrot
& 20.0 & 90.0 & 90.0
& 10.0 & 40.0 & 70.0
& 20.0 & 40.0 & 50.0 \\

PnP car
& 40.0 & 100.0 & 100.0
& 0.0 & 90.0 & 100.0
& 30.0 & 60.0 & 100.0 \\

PnP bread
& 40.0 & 90.0 & 100.0
& 20.0 & 40.0 & 100.0
& 40.0 & 90.0 & 100.0 \\

PnP carrot
& 50.0 & 90.0 & 100.0
& 10.0 & 70.0 & 70.0
& 30.0 & 80.0 & 70.0 \\

PnP egg
& 20.0 & 70.0 & 70.0
& 10.0 & 20.0 & 60.0
& 30.0 & 50.0 & 60.0 \\

PnP pear
& 10.0 & 70.0 & 60.0
& 0.0 & 30.0 & 50.0
& 0.0 & 70.0 & 70.0 \\

Sort toys
& 50.0 & 80.0 & 100.0
& 0.0 & 10.0 & 80.0
& 30.0 & 90.0 & 90.0 \\

Wipe table
& 50.0 & 80.0 & 100.0
& 0.0 & 60.0 & 100.0
& 50.0 & 100.0 & 100.0 \\

\midrule
\textbf{Average}
& 30.0 & \underline{77.3} & \textbf{87.3}
& 7.3 & \underline{47.3} & \textbf{74.5}
& 23.6 & \underline{66.4} & \textbf{80.9} \\

\bottomrule
\end{tabular*}

\endgroup

\caption{$\pi_0$, Fast-WAM, and JEPA-WAM success rates (\%) under ID, OOD-S, and OOD-I
settings in real robot rollouts. JEPA-WAM uses a 1,024 visual-instruction bank.
The best and second-best results are shown in bold and underlined,
respectively.}
\label{tab:main_results}
\end{table*}

\subsection{Evaluation Protocol and Metrics}
\label{sec:benchmark_protocol}

Each task is evaluated over ten real-robot rollouts in every setting, yielding
110 rollouts per setting and 330 rollouts per policy. Five trials place the box
on the left and the plate on the right, while the remaining five reverse this
arrangement. Object positions are randomized within the workspace, but every
evaluated policy receives the same set of ten predefined initial configurations
for each task and setting.

A separate operator loads and launches each policy, while the evaluator who
conducts, judges, and records the rollouts is not informed of the policy
identity. No human intervention is permitted after a rollout begins. We use
binary task success and award no partial credit. For \textit{Give Me}, an
object of the instructed category must touch or cross the front table boundary.
For PnP, the target object must be fully placed in or on the correct container
and released by the gripper. For \textit{Sort Toys}, all food and non-food
objects must be assigned to their corresponding containers. For \textit{Wipe Table}, the robot must wipe the exposed tabletop
regions and remove the designated clutter objects from the table,
while leaving the box, plate, and objects near the table boundary
undisturbed. The towel must then be returned to its basket. Partial completion, incorrect
outcomes, behavior switching, and unrecoverable object drops are
counted as failures.

We report the success rate over ten rollouts for each task--setting
pair and the mean across all 11 tasks. Since every task has the same
number of trials, the task-level mean equals the overall success rate
over the 110 rollouts in each setting.

\section{Experiments}
\label{sec:experiments}
We organize our experiments around four questions: \textbf{(Q1)} Does JEPA-WAM
improve instruction following under familiar scenes and instructions?
\textbf{(Q2)} Does it remain robust to scene and instruction shifts that expose
reliance on visual or linguistic shortcuts? \textbf{(Q3)} How does performance
change as the generated visual-instruction bank scales? \textbf{(Q4)} Are
generated visual instructions more effective than real demonstration
keyframes? We first describe the experimental setup and then address these
questions through the main comparison and two controlled ablations.
\subsection{Experimental Setup}
\label{sec:experimental_setup}

\begin{table*}[!t]
\centering

\begingroup
\setlength{\tabcolsep}{0.7mm}

\begin{tabular*}{\textwidth}{
    @{\extracolsep{\fill}}
    l
    cccc
    cccc
    cccc
    @{}
}
\toprule
& \multicolumn{4}{c}{ID}
& \multicolumn{4}{c}{OOD-S}
& \multicolumn{4}{c}{OOD-I} \\
\cmidrule(lr){2-5}
\cmidrule(lr){6-9}
\cmidrule(lr){10-13}

Task
& KF-64 & J-64 & J-512 & J-1024
& KF-64 & J-64 & J-512 & J-1024
& KF-64 & J-64 & J-512 & J-1024 \\
\midrule

Give bread
& 100.0 & 80.0 & 70.0 & 100.0
& 60.0 & 70.0 & 60.0 & 90.0
& 70.0 & 80.0 & 70.0 & 100.0 \\

Give pear
& 50.0 & 60.0 & 70.0 & 60.0
& 40.0 & 50.0 & 60.0 & 50.0
& 60.0 & 70.0 & 70.0 & 90.0 \\

Give egg
& 70.0 & 70.0 & 80.0 & 80.0
& 30.0 & 60.0 & 60.0 & 50.0
& 40.0 & 40.0 & 60.0 & 60.0 \\

Give carrot
& 60.0 & 80.0 & 70.0 & 90.0
& 50.0 & 50.0 & 70.0 & 70.0
& 10.0 & 40.0 & 80.0 & 50.0 \\

PnP bread
& 90.0 & 90.0 & 100.0 & 100.0
& 100.0 & 80.0 & 80.0 & 100.0
& 100.0 & 80.0 & 90.0 & 100.0 \\

PnP carrot
& 60.0 & 90.0 & 80.0 & 100.0
& 60.0 & 80.0 & 60.0 & 70.0
& 90.0 & 60.0 & 70.0 & 70.0 \\

PnP egg
& 40.0 & 60.0 & 60.0 & 70.0
& 20.0 & 40.0 & 50.0 & 60.0
& 30.0 & 50.0 & 60.0 & 60.0 \\

PnP pear
& 80.0 & 60.0 & 70.0 & 60.0
& 20.0 & 60.0 & 60.0 & 50.0
& 30.0 & 50.0 & 60.0 & 70.0 \\

\midrule
\textbf{Average}
& 68.8 & 73.8 & \underline{75.0} & \textbf{82.5}
& 47.5 & 61.2 & \underline{62.5} & \textbf{67.5}
& 53.8 & 58.8 & \underline{70.0} & \textbf{75.0} \\

\bottomrule
\end{tabular*}

\endgroup

\caption{Success rates (\%) for visual-instruction banks constructed
from real demonstration keyframes or generated images on the eight shared
tasks. KF-64 denotes a bank of 64 real demonstration keyframes, while
J-$K$ denotes a generated visual-instruction bank containing $K$ images.
The highest and second-highest average under each evaluation setting are
shown in bold and underlined, respectively.}
\label{tab:visual_bank_ablation}
\end{table*}

We compare JEPA-WAM with two representative robot-policy baselines.
$\pi_0$ is initialized from its officially released pretrained checkpoint.
Fast-WAM is initialized from a checkpoint pretrained on approximately 5,000
hours of heterogeneous real-robot manipulation data. JEPA-WAM uses the same
pretrained Fast-WAM checkpoint, together with a newly initialized JEPA
projector; the DreamLite generator and V-JEPA 2.1 encoder remain frozen. The
1,370 benchmark demonstrations were collected separately and were not used in
the pretraining of any evaluated model. The Fast-WAM comparison therefore isolates the contribution of the
generated visual-instruction pathway, whereas $\pi_0$ provides a representative
VLA baseline with its native pretrained initialization.

All methods are fine-tuned on the same 1,370 demonstrations using the same
per-process batch size and distributed hardware. Fast-WAM and JEPA-WAM are evaluated at the matched checkpoint. Because $\pi_0$ showed degraded closed-loop behavior at this checkpoint, we selected its best checkpoint among 6k, 9k, and 12k steps using a disjoint pilot set; the 9k checkpoint was fixed before full evaluation, and pilot rollouts were excluded. Full training details are provided in
Supplementary Sec.~C.

In total, the reported experiments comprise 1,710 distinct real-robot
rollouts. All methods are evaluated using the protocol described in
Sec.~\ref{sec:benchmark_protocol}.

\subsection{Main Instruction-Following Results}
\label{sec:main_results}

Table~\ref{tab:main_results} reports the complete 11-task benchmark results for
$\pi_0$, Fast-WAM, and JEPA-WAM under ID, OOD-S, and OOD-I.

\paragraph{In-distribution instruction following.}
JEPA-WAM achieves an average success rate of 87.3\% under ID, compared with
77.3\% for Fast-WAM and 30.0\% for $\pi_0$. It matches or exceeds Fast-WAM on
nine of the 11 tasks, showing that visual-instruction conditioning improves
task disambiguation without reducing execution performance in familiar scenes.
Qualitatively, WAM-based failures primarily involve confusion among visually
similar objects, whereas $\pi_0$ also frequently confuses handover and
placement behaviors.

\paragraph{Robustness to scene shifts.}
Under OOD-S, JEPA-WAM achieves 74.5\% success, outperforming Fast-WAM at 47.3\%
and $\pi_0$ at 7.3\%. The improvement is particularly pronounced on Sort Toys,
where JEPA-WAM reaches 80.0\% despite the introduction of object categories
absent from the sorting demonstrations, compared with 10.0\% for Fast-WAM and
0.0\% for $\pi_0$. This result indicates reduced reliance on familiar visual
configurations when multiple plausible objects and behaviors are present.

\paragraph{Robustness to instruction shifts.}
Under OOD-I, JEPA-WAM achieves 80.9\% success, compared with 66.4\% for
Fast-WAM and 23.6\% for $\pi_0$. Its gains are largest when unseen
reformulations resemble instructions for behaviorally related tasks, including
40-point improvements over Fast-WAM on Give Bread and PnP Car. These results
suggest that generated visual instructions help preserve task semantics under
lexical and structural shifts.

\subsection{Goal-Bank Scaling}
\label{sec:goal_bank_scaling}

We compare visual-instruction banks containing 64, 512, and 1,024 generated
images while fixing the pretrained WAM initialization, architecture,
demonstrations, training budget, and evaluation protocol. The comparison uses
the four Give Me tasks and four food PnP tasks shared by all variants.

As shown in Table~\ref{tab:visual_bank_ablation}, scaling the bank from 64 to 1,024
images improves average success from 73.8\% to 82.5\% under ID, from 61.2\% to
67.5\% under OOD-S, and from 58.8\% to 75.0\% under OOD-I. The effect of
scaling differs across settings. Increasing the bank from 64 to 512 produces
modest gains of 1.2 and 1.3 percentage points under ID and OOD-S, respectively,
but improves OOD-I by 11.2 points. Increasing it further to 1,024 yields the
highest average performance in all three settings. Although individual tasks
do not improve monotonically, the average results support the benefit of
broader visual variation without additional robot demonstrations or
optimization steps.

\subsection{Generated Visual Instructions vs. Real Keyframes}
\label{sec:generated_vs_keyframes}

We compare JEPA-WAM-64 with Keyframe-64, which replaces generated images with
64 manually verified frames from real demonstrations. Both variants use the
same pretrained WAM initialization, V-JEPA encoder, projector architecture,
demonstrations, bank size, and training budget. Construction details are
provided in Supplementary Sec.~D.

As shown in Table~\ref{tab:visual_bank_ablation}, generated visual instructions achieve higher average success in all three
settings: 73.8\% versus 68.8\% under ID, 61.2\% versus 47.5\% under OOD-S, and
58.8\% versus 53.8\% under OOD-I. This result is notable because the real
keyframes are task-relevant future states selected from successful robot
demonstrations. They therefore provide physically valid references that match
the robot embodiment, camera viewpoint, and workspace appearance, without
image-generation artifacts.

One possible explanation is that generated images tend to place the manipulated
objects and task-defining relations more prominently, whereas top-camera
keyframes allocate more of the image to the surrounding workspace. Generated
references may therefore provide a denser task-semantic signal to the JEPA
encoder. In addition, their stochastic variation spans a wider range of appearances and
layouts than demonstration keyframes tied to the training workspace. This may
reduce dependence on training-specific visual context, which is consistent
with the larger improvement under OOD-S.

\section{Conclusion}

We introduced JEPA-WAM, which connects stochastically generated visual
instructions to World Action Models through JEPA latent representations. On our
11-task real-robot benchmark, JEPA-WAM consistently improves instruction
following under ID, OOD-S, and OOD-I conditions, with the largest gain occurring
under scene shifts. Controlled ablations further demonstrate the benefits of
scaling the visual-instruction bank and using generated images instead of real
demonstration keyframes. These results support generated visual instructions
as a scalable source of task-semantic conditioning for improving instruction-following robustness of World Action Models under scene and instruction shifts in behaviorally overlapping tabletop manipulation tasks.

\appendix

\bibliography{aaai2027}


\clearpage
\appendix

\section*{Supplementary Material}

\section{Image-Generation Configuration}
\label{app:generation_prompt}

We use the DreamLite-base checkpoint with the \texttt{diffusers} revision for
all generated visual-instruction banks. Images are generated at a resolution
of $768\times768$. We use the following fixed generation prompt:

\begin{quote}
\small
\texttt{a dual-arm robot performing task: \{task\}. Generate the moment the task is completed.
}
\end{quote}

Here, \texttt{\{task\}} is replaced by the language instruction associated
with the corresponding robot task. The same template is used for every task,
without task-specific prompt tuning.

For a visual-instruction bank of size $B$, we generate one image for each seed
in $\{42,43,\ldots,42+B-1\}$. We evaluate bank sizes of $B\in
\{64,512,1024\}$, corresponding to 64, 512, or 1,024 generated images for each
instruction. Unless otherwise specified, the main JEPA-WAM model uses
$B=1024$. During training, an image is sampled uniformly from the bank
associated with the current instruction.

\section{Complete scene compositions of Robot Instruction-Following Benchmark}
In the \textit{Give Me} tasks, the robot moves an object of the instructed
category to the front handover boundary of the table. For \textit{Give Bread},
the workspace contains two toy cars and three pieces of bread in addition to
the box and plate. The other three variants, \textit{Give Pear},
\textit{Give Egg}, and \textit{Give Carrot}, share the same
scene configuration, containing two toy
cars and one instance each of pear, egg, and carrot. These tasks use
instructions of the form ``give me the [object].''

In the PnP tasks, the robot places a toy car in the box or places an edible
object---bread, pear, egg, or carrot---on the plate. The \textit{PnP Car} and
\textit{PnP Bread} variants share the same
scene configuration, containing two toy cars and three pieces of
bread. The \textit{PnP Pear}, \textit{PnP Egg}, and \textit{PnP Carrot}
variants share the same
scene configuration, containing a toy car and one instance each of pear, egg,
and carrot.

The \textit{Sort Toys} demonstrations contain two toy cars and three
pieces of bread. The robot first places both cars in the box and then
places all three pieces of bread on the plate. These are the only two
object categories used to demonstrate the sorting behavior during
training.

The \textit{Wipe Table} demonstrations require the robot to retrieve a towel
from a basket beside the tabletop, wipe the unoccupied tabletop regions while
avoiding containers and other objects, and return the towel to its original
basket. In addition to the fixed scene elements, one food object is placed near
the table edge with probability $0.25$, two food objects are present with
probability $0.25$, and no additional food object is present with probability $0.5$.

\paragraph{OOD-S Setting.} 
This setting preserves the training instructions while introducing
distractor-rich scenes and unseen object categories. For the \textit{Give Me}
and PnP tasks, the workspace contains all six object categories appearing in
the demonstration dataset---car, bread, pear, egg, carrot, and towel---together
with three previously unseen objects: one toy component, a chili
pepper, and a green onion. The policy must select the instructed object and
execute the requested behavior among multiple graspable objects and plausible
manipulation alternatives.

For \textit{Sort Toys}, the towel is excluded because it is not a sorting
object. The evaluation scene contains car, bread, pear, egg, and carrot,
together with the unseen toy component, chili pepper, and green onion.
Although the sorting demonstrations contain only the car and bread categories, the policy
must place all non-food objects in the box and all food objects on the
plate. This setting tests whether the demonstrated category-to-container
relation transfers to categories absent from the sorting demonstrations,
including objects absent from the benchmark training data.

For Wipe Table, four OOD-S trials contain familiar removable clutter, three
contain unseen removable clutter, and three contain a mixture of both. The
policy must clear this clutter while wiping the exposed tabletop, without
disturbing the box, plate, or protected food objects placed near the table
boundary, and must finally return the towel to its basket.

\paragraph{OOD-I Setting.}
This setting retains an ID-like scene and the intended task while replacing
the training instruction with a semantically equivalent but previously unseen
formulation. For each task, we manually define one OOD-I instruction before
evaluation and use it unchanged across all ten trials and all evaluated
policies. Some OOD-I instructions are deliberately designed to overlap
lexically and structurally with instructions for behaviorally related tasks.
For example, ``give me the egg'' is reformulated as ``pick up the egg and pass
it to me,'' which resembles the PnP instruction ``pick up the egg and place it
on the plate'' while specifying a different outcome. The remaining instructions
use more direct semantic paraphrases, such as reformulating ``give me the
bread'' as ``bring me the bread.'' OOD-I therefore tests both robustness to
unfamiliar linguistic formulations and the ability to follow complete
instruction semantics rather than matching familiar keywords or sentence
templates. The complete list of ID and OOD-I instruction pairs is provided in Table~\ref{tab:ood_instructions} . The OOD-I instructions preserve the intended behavior while
using different lexical or syntactic forms, several of which resemble
instructions for behaviorally related tasks.

\begin{table*}[!t]
\centering
\small
\setlength{\tabcolsep}{4pt}
\renewcommand{\arraystretch}{1.1}

\begin{tabular}{
    @{}
    p{0.14\textwidth}
    p{0.38\textwidth}
    p{0.38\textwidth}
    @{}
}
\toprule
Task & ID instruction & OOD-I instruction \\
\midrule

Give bread
& ``give me the bread''
& ``bring me the bread'' \\

Give pear
& ``give me the pear''
& ``hand me the pear'' \\

Give egg
& ``give me the egg''
& ``pick up the egg and pass it to me.'' \\

Give carrot
& ``give me the carrot''
& ``pick up the carrot and give it to me.'' \\

PnP car
& ``pick up the toy car and place it into the box''
& ``put the toy car into the box.'' \\

PnP bread
& ``pick up the bread and place it on the plate''
& ``put the bread onto the plate.'' \\

PnP carrot
& ``pick up the carrot and place it on the plate''
& ``place the carrot onto the plate'' \\

PnP egg
& ``pick up the egg and place it on the plate''
& ``place the egg onto the plate'' \\

PnP pear
& ``pick up the pear and place it on the plate''
& ``put the pear onto the plate'' \\

Sort toys
& ``Sort toys on the table.''
& ``Sort toys into the containers.'' \\

Wipe table
& ``use the towel to wipe the table''
& ``wipe the table with the towel.'' \\

\bottomrule
\end{tabular}

\caption{Instructions used under ID and OOD-I evaluation. OOD-I instructions
are not present in the robot demonstration dataset.}
\label{tab:ood_instructions}
\end{table*}

\begin{figure*}[!t]
    \centering
    \includegraphics[width=\textwidth]{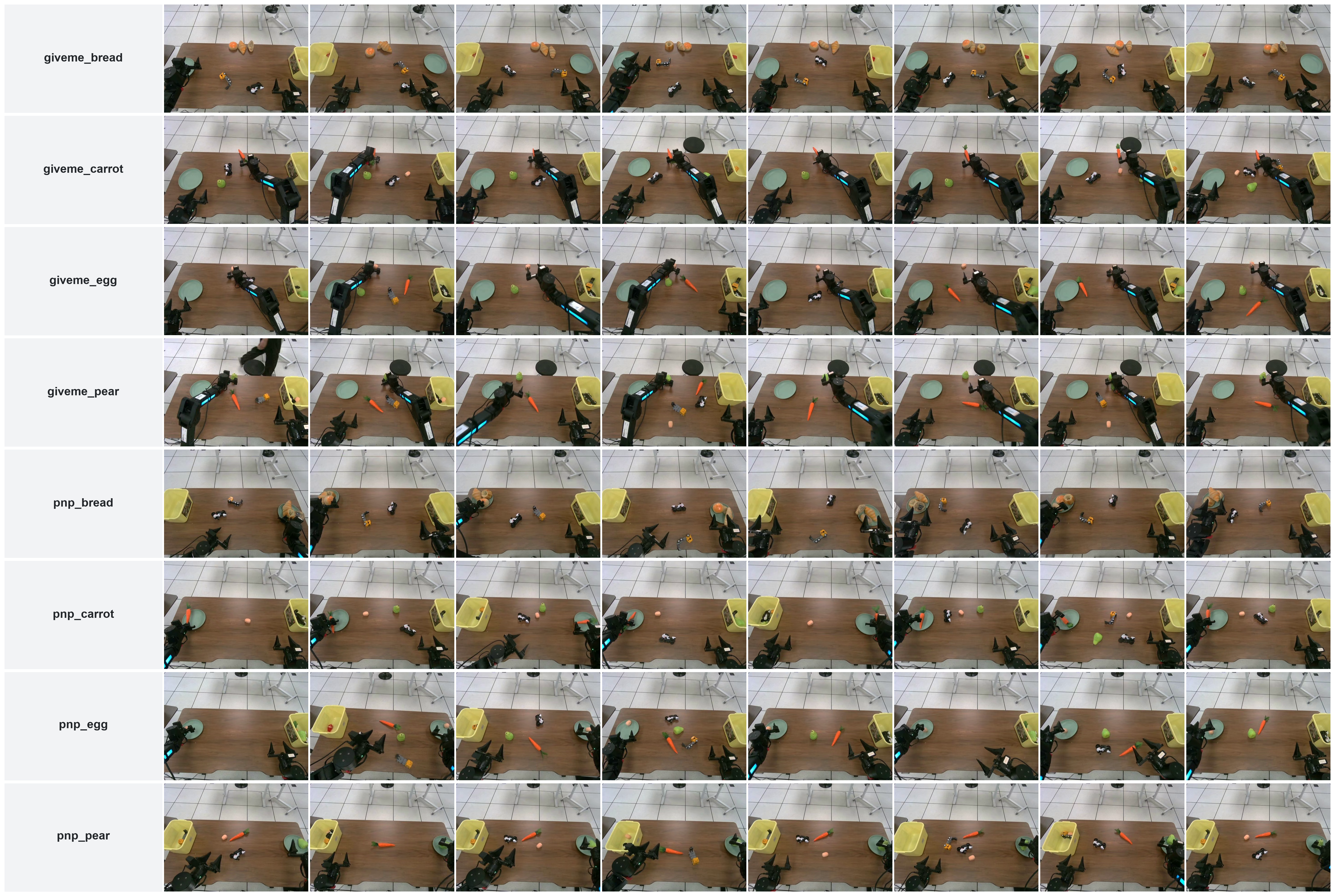}
    \caption{Representative real keyframes retained for the Keyframe-64
    baseline. Each frame is manually selected from a distinct training
    demonstration and retained only when the task-relevant objects and
    intended interaction or outcome are clearly visible.}
    \label{fig:keyframe_examples}
\end{figure*}

\section{Training and Implementation Details}
\label{app:train_details}

\paragraph{Model initialization.}
Fast-WAM is initialized from a checkpoint pretrained on a 5,000-hour corpus.
All Fast-WAM and JEPA-WAM variants use this identical checkpoint. JEPA-WAM
additionally introduces a randomly initialized linear projector, which is
optimized jointly with the WAM during benchmark fine-tuning. The DreamLite
generator and V-JEPA 2.1 encoder remain frozen. The $\pi_0$ baseline is
initialized from the officially released pretrained checkpoint.

\paragraph{Benchmark fine-tuning.}
All models are fine-tuned on the same 1,370 demonstrations using 72 NVIDIA H20
GPUs, a per-process batch size of 16, gradient accumulation of one, and
bfloat16 precision, resulting in an effective global batch size of 1,152.

For Fast-WAM and all JEPA-WAM variants, we optimize the WAM parameters with
AdamW using $\beta_1=0.9$, $\beta_2=0.95$, a learning rate of
$2\times10^{-5}$, weight decay of $10^{-2}$, gradient clipping at $1.0$, and a
cosine learning-rate schedule with a $5\%$ warmup period. In JEPA-WAM, the
newly initialized JEPA projector is optimized jointly with the WAM, while the
DreamLite generator and V-JEPA 2.1 encoder remain frozen.

Following its official fine-tuning implementation, $\pi_0$ is initialized from
the officially released pretrained checkpoint and only its action head is
fine-tuned. Input images are resized to $224\times224$, and the policy predicts
action chunks of 50 steps. We use AdamW with $\beta_1=0.9$, $\beta_2=0.98$,
$\epsilon=10^{-7}$, a learning rate of $2\times10^{-5}$, zero weight decay,
and a cosine learning-rate schedule with a $1\%$ warmup period. TF32 is
enabled, while relative action control and reasoning-augmented instructions
are disabled.

\paragraph{Checkpoint selection.}
Fast-WAM and all JEPA-WAM variants are evaluated at the matched 17,880-step
checkpoint. At 17,880 steps, $\pi_0$ exhibits behavioral collapse, producing
nearly the same behavior for different instructions. To avoid underestimating the $\pi_0$ baseline, we therefore conduct a
preliminary evaluation of its 6,000-, 9,000-, and 12,000-step checkpoints and
select the 9,000-step checkpoint, which is fixed before the full benchmark
evaluation. The pilot rollouts are disjoint from and excluded from the reported
evaluation results.

\paragraph{Inference.}
Real-robot inference is performed on a single NVIDIA RTX 5090. The WAM-based
policies use ten action-denoising steps. For JEPA-WAM, DreamLite generates a
visual instruction when a new language instruction is received. The generated
image and its V-JEPA representation are reused until the language instruction
changes, avoiding repeated image generation during closed-loop execution.

\section{Real-Keyframe Construction}
\label{app:keyframe}

The Keyframe-64 banks are constructed exclusively from the benchmark training
demonstrations; no pilot or evaluation rollout is used. For each instruction,
we manually inspect its demonstration episodes and select at most one frame
from each episode. The selected frame is the one that most clearly expresses
the task semantics, including the task-relevant objects and the intended
interaction or outcome.

We discard an episode if none of its frames provides an unambiguous visual
description of the task, or if the manipulated object and interaction are
occluded by the robot arms or otherwise not visible from the primary camera.
We then continue inspecting subsequent episodes until 64 valid keyframes have
been collected for the instruction. Because every task contains at least 80
demonstrations, this procedure produces a 64-image bank for each evaluated
instruction despite the rejected episodes.

During training, one keyframe is sampled uniformly from the bank associated
with the current instruction. The keyframes use the same image preprocessing,
frozen V-JEPA 2.1 encoder, and latent-projector architecture as the generated
visual instructions. Thus, Keyframe-64 and JEPA-WAM-64 differ only in the
source of their visual-reference banks. Figure~\ref{fig:keyframe_examples}
shows representative keyframes retained by this procedure.

\end{document}